\documentclass[runningheads]{llncs}

\usepackage[T1]{fontenc}
\usepackage{fmtcount}
\usepackage[colorlinks,linkcolor=blue]{hyperref}
\usepackage{graphicx,verbatim}
\usepackage{amssymb}
\usepackage{amsmath}
\usepackage{amssymb}
\usepackage{multirow}
\usepackage{tabularx}

\usepackage{url}
\usepackage[table,xcdraw]{xcolor}
\usepackage{amsmath}
\usepackage{xcolor}
\usepackage{graphicx} 
\usepackage{booktabs} 
\usepackage[misc]{ifsym}

\begin{document}
\title{Hierarchical Corpus-View-Category Refinement for Carotid Plaque Risk Grading in Ultrasound}

\author{
Zhiyuan Zhu\inst{1,2}\thanks{Zhiyuan Zhu, Jian Wang and Yong Jiang contribute equally to this work.\\Corresponding email: \email{821473120@qq.com} and  
\email{xinyang@szu.edu.cn}}\and 
Jian Wang\inst{1,2,3\star} \and 
Yong Jiang\inst{4\star} \and    
Tong Han\inst{1,2} \and         
Yuhao Huang\inst{1,2} \and      
Ang Zhang\inst{5} \and          
Kaiwen Yang\inst{6} \and        
Mingyuan Luo\inst{1,2} \and     
Zhe Liu\inst{1,2} \and          
Yaofei Duan\inst{6} \and        
Dong Ni\inst{1,2,3} \and            
Tianhong Tang\inst{4}\textsuperscript{(\Letter)} \and  
Xin Yang\inst{1,2}\textsuperscript{(\Letter)}       
} 

\authorrunning{Zhu et al.}
\titlerunning{CVC-RF}

\institute{
National-Regional Key Technology Engineering Laboratory for Medical Ultrasound, School of Biomedical Engineering, Medical School, Shenzhen University, China \and 
Medical Ultrasound lmage Computing (MUSIC) Lab, School of Biomedical Engineering, Medical School, Shenzhen University, China \and
College of Computer Science and Software Engineering, Shenzhen University, China \and
Fuwai Shenzhen Hospital, Chinese Academy of Medical Sciences, Shenzhen, China \and
Shenzhen RayShape Medical Technology Co., Ltd, China \and 
Faculty of Applied Sciences, Macao Polytechnic University, China
}
\maketitle

\begin{abstract}
Accurate carotid plaque grading (CPG) is vital to assess the risk of cardiovascular and cerebrovascular diseases. Due to the small size and high intra-class variability of plaque, CPG is commonly evaluated using a combination of transverse and longitudinal ultrasound views in clinical practice. However, most existing deep learning-based multi-view classification methods focus on feature fusion across different views, neglecting the importance of representation learning and the difference in class features. To address these issues, we propose a novel Corpus-View-Category Refinement Framework (CVC-RF) that processes information from Corpus-, View-, and  Category-levels, enhancing model performance. Our contribution is four-fold. First, to the best of our knowledge, we are the foremost deep-learning-based method for CPG according to the latest Carotid Plaque-RADS guidelines. Second, we propose a novel center-memory contrastive loss, which enhances the network's global modeling capability by comparing with representative cluster centers and diverse negative samples at \textbf{Corpus-level}. Third, we design a cascaded down-sampling attention module to fuse multi-scale information and achieve implicit feature interaction at \textbf{View-level}. Finally, a parameter-free mixture-of-experts weighting strategy is introduced to leverage class clustering knowledge to weight different experts, enabling feature decoupling at \textbf{Category-level}. Experimental results indicate that CVC-RF effectively models global features via multi-level refinement, achieving state-of-the-art performance in the challenging CPG task.The source is available at \href{https://github.com/dndins/CVC-RF}{https://github.com/dndins/CVC-RF}
\end{abstract}
\begin{figure}[htbp]
    \centering
    \includegraphics[width=1\textwidth]{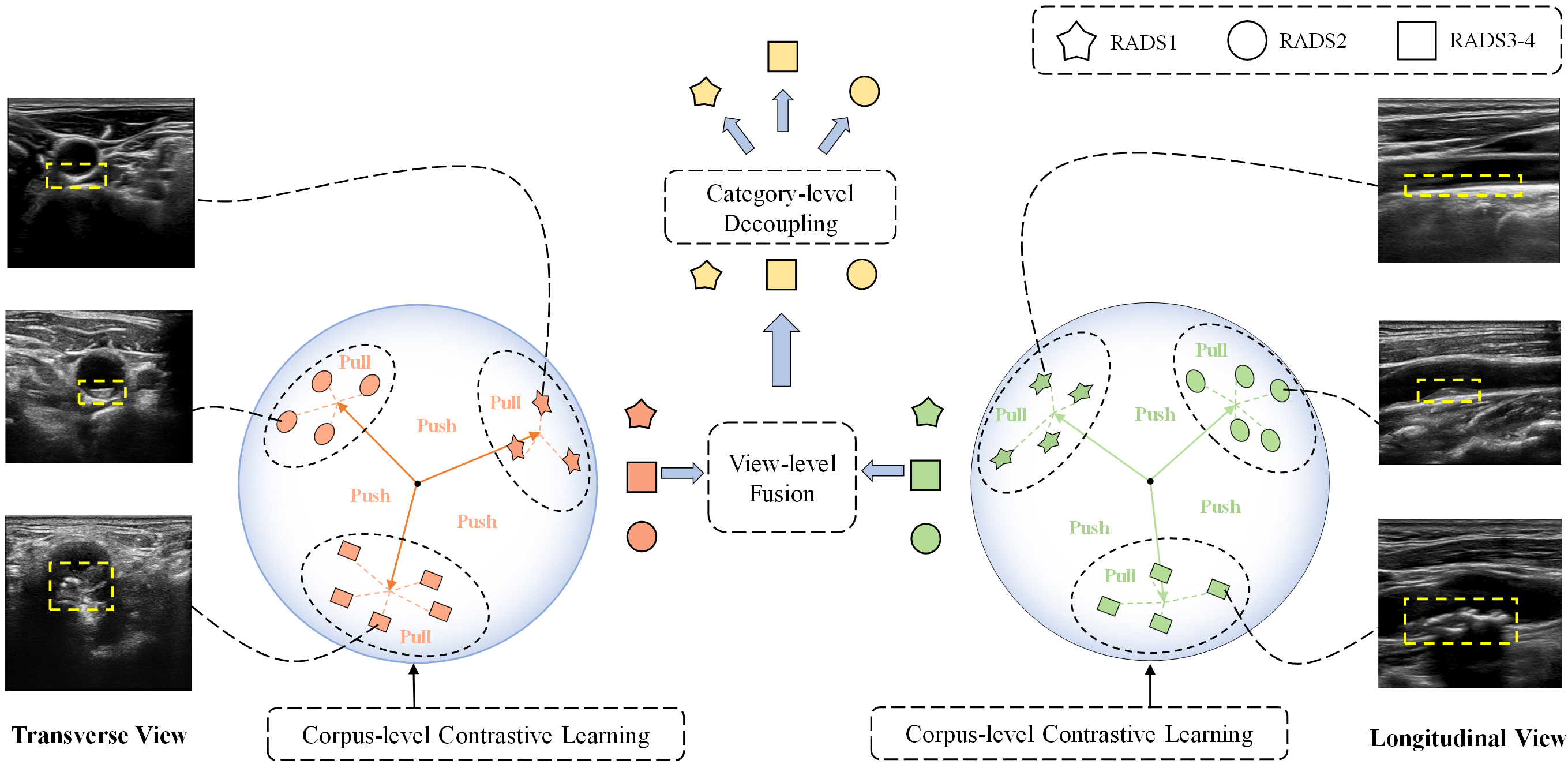}
    \caption{
    The information refinement flowchart includes Corpus-level contrastive learning for better representation learning, View-level fusion to extract shared information, and Category-level decoupling for class-specific learning. The yellow dotted boxes indicate the vessel wall or plaque.
    }
    \label{fig1}
\end{figure}
\section{Introduction}
Acute ischemic stroke has become the second leading cause of death in the world, high-risk carotid plaque rupture is one of the main contributors to embolism-induced stroke \cite{r1}. 
Accurate carotid plaque grading (CPG) is crucial for early diagnosis and prevention of stroke. 
Ultrasound (US) as a wide availability, real-time imaging modality, has become the primary
technique used in CPG. However, due to complex characteristics and the small size of plaques~\cite{r10}, CPG demands extensive expertise, making it challenging for less-experienced clinicians.

To alleviate this problem, several deep learning(DL)-based methods have been adopted for CPG~\cite{r2,r3,r4}. For example, Zhang et al.~\cite{r6} proposed a KNN classifier to perform binary classification of carotid plaques. Singh et al. \cite{r7} compared the performance of CNN-based models in classifying plaque levels in US images. Zhou et al. \cite{r9} used a hybrid Convolution-Transformer approach on US videos to assess the degree of carotid stenosis. 
Although effective, the above studies focus primarily on one single view, i.e., transverse or longitudinal; however, clinical carotid US scans are typically performed in both views for complete diagnostic information. As plaque has large intra-class variation~\cite{r10}, single-view imaging may not provide sufficient information for accurate CPG. These limitations highlight the need to combine multi-view learning to improve diagnostic accuracy and reliability.

Most recently, integrating multi-view information for disease diagnosis has demonstrated significant advantages. Huang et al. \cite{Huang} combined transverse and longitudinal thyroid US images to classify benign and malignant tumors. Luo et al. \cite{r12} aligned sagittal, coronal, and axial CT scans to enhance lung cancer histological classification. Huang et al. \cite{r13} introduced a reinforcement learning method to weight four-modal US breast tumor images, exploring the dynamic integration of information from different modalities. However, these methods focus primarily on extracting high-dimensional features from different views for fusion or applying decision weighting to each view, neglecting the importance of representation learning and inter-class feature differences. Moreover, they lack specific design considerations for small structures, increasing the risk of information loss and performance degradation.

In this study, we propose a Corpus-View-Category Refinement Framework (CVC-RF) to address these problems. As illustrated in Fig.~\ref{fig1}, we decompose the multi-view classification task into three distinct information dimensions: \textbf{Corpus-level}, \textbf{View-level} and \textbf{Category-level}. By hierarchical refinement information across three levels, it can effectively enhance the model's feature learning capacity, thereby improving its overall performance. We refer to the latest carotid plaque-RADS guidelines~\cite{RADS} for CPG, categorizing plaques into RADS1, RADS2, and RADS3-4. Our contributions can be summarized as follows: \textbf{1)} We present the first DL-based work to follow the plaque-RADS guidelines for CPG. \textbf{2)} We propose a novel Center Memory Contrastive Loss (CMCL) that leverages positive class centers from a memory bank alongside diversified negative samples to calculate the loss, constraining intra-class aggregation and inter-class separation, guiding the model to learn global features distribution. \textbf{3)} A cascaded down-sampling attention module (DSAM) that progressively fuses multi-scale features, enhancing cross-view information interaction and improving the capture of small-sized plaque details. \textbf{4)} A parameter-free mixture-of-experts (MoE) weighting strategy for category feature decoupling, allowing each expert to focus on specific features without extra parameters.
Experimental results confirm that CVC-RF surpasses current approaches, validating our framework's efficacy.
\section{Methods}
\begin{figure}[htbp]
    \centering
    \includegraphics[width=1\textwidth]{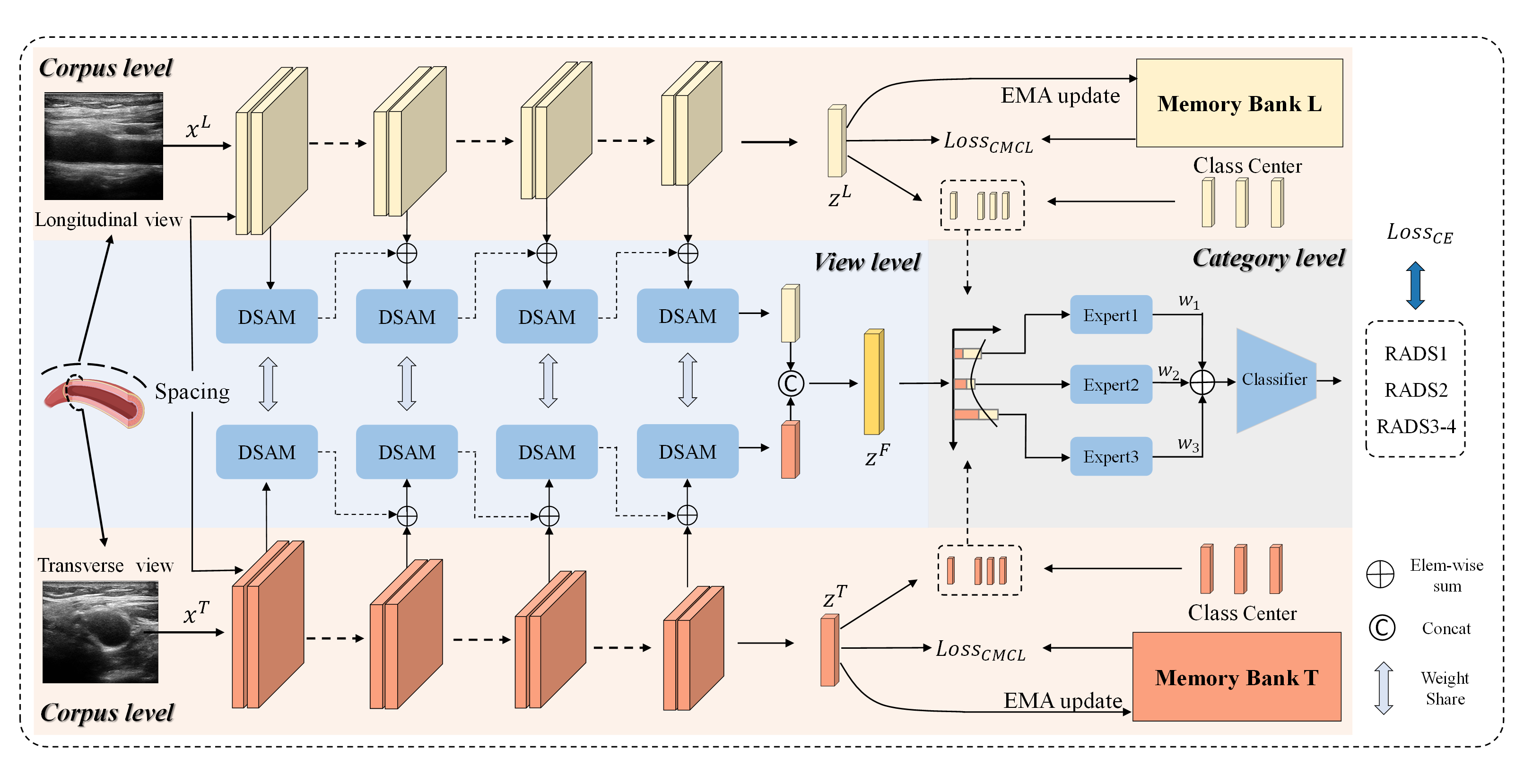}
    \caption{Overview of our proposed framework}
    \label{fig2}
\end{figure}

Our CVC-RF is shown in Fig.~\ref{fig2}. 
The longitudinal and transverse carotid images of one patient are fed into the ResNet18 \cite{ResNet18} backbone to extract intermediate and representation features. In this step, considering that plaque thickness is a crucial clinical indicator in CPG, we reshaped the spacing ( physical size of each pixel) to image size and concatenated it in channel dim with images, This combined input was then fed into CVC-RF to ensure alignment between pixel space and physical space, allowing the network to perceive plaque size. The representation features then interact with the corresponding memory bank to compute the CMCL and acquire the weights $w_{i}$ for each MoE expert. Before training, all images are encoded by a ResNet-18 (ImageNet-pretrained) and projected into 128-dim features, then stored in the memory bank for initialization. On the other hand, the intermediate features are processed through a cascaded DSAM to obtain the multi-scale fusion features from two different views. Finally, these fusion features are concatenated and fed into MoE. Each expert's output is summed based on the obtained $w_{i}$ and passed to the classifier to obtain the final prediction.
\subsection{Memory Bank-based Center-memory Contrastive Loss}
\label{2.1}
Previous contrastive learning methods~\cite{moco,simclr} perform positive and negative comparisons using only samples within a large batch to better approximate the global distribution. However, some bias inevitably remains unless the entire training set is used as a single batch. In this part, we introduce the CMCL, which allows features of the same class to cluster toward a stable center in the feature space while pushing apart features from different classes globally, helping the model establish a clear decision boundary and learning an unbiased global feature distribution.

Let $\mathcal{D}_{\text{train}}=\{x^L_i, x^T_i, y_i\}_{i=1}^{N}$ be the training set of images. $y_{i}\in\{1, 2, \dots, K\}$, $K$ is the number of classes. $x_{i}^{L}$ and $x_{i}^{T}$ are longitudinal and transverse images of one patient. $z_{i}^{L}$ and $z_{i}^{T}$ are the representation features obtained from $x_{i}^{L}$ and $x_{i}^{T}$ through the encoder $f(\cdot)$, then $z_{i}^{L}$, $z_{i}^{T}$ and the label $y_{i}$ are used to update the memory bank $\mathcal{M}=\{m^L_j, m^T_j, y_j\}_{j=1}^{N}$. The elements in $\mathcal{D}_{\text{train}}$ and $\mathcal{M}$ correspond one-to-one. Since both views undergo the same operations, for the sake of clarity, we describe only the longitudinal view, with the other view processed identically.

For every training epoch $t$, the representation feature $z_{i}^{L}$ serves three purposes. 
\textbf{First}, they update $\mathcal{M}$ using an exponential moving average (EMA) strategy, ensuring that $\mathcal{M}$ adapts to the change of the model as its parameters are optimized. The update process is as follows: 
\begin{equation}
m_{j,t+1}^{L}\gets \alpha m_{j,t}^{L} +(1-\alpha)z_{i}^{L},
\label{eq:EMA}
\end{equation}
where $\alpha \in $[0, 1] is the momentum coefficient that controls the weight update speed.
\textbf{Second}, $z_{i}^{L}$ interacts with all class centers in $\mathcal{M}$ to calculate cosine similarity. For any class $k$, the corresponding class center $\mu_{k}^{L}$ is defined as: 
\begin{equation}
\mu_{k}^{L} = \frac{1}{\left | C_{k} \right | } \sum_{j\in C_{k}}^{} m_{j}^{L},
\label{eq:class center}
\end{equation}
where $C_k = \{j \mid y_j = k\}$ represents as the set of all samples in $\mathcal{M}$ that belonging to class $k$. \textbf{Third}, $z_{i}^{L}$ is used to compute the CMCL, which updates the parameters of encoder $f(\cdot)$. For the feature $z_{i}^{L}$ extracted from $x_{i}^{L}$, it corresponds to the label $y_{i} = k$. The positive pair for $z_{i}^{L}$ is the class center $u_{k}^{L}$, while the negative pairs are all feature samples in $\mathcal{M}$ that do not belong to class $k$. The loss corresponding to $z_{i}^{L}$ is given by:
\begin{equation}
     L_{CMCL}^{L} = - \log\frac{\exp(sim(z_{i}^{L}, \mu_{k}^{L})/\tau)}{\exp(sim(z_{i}^{L}, \mu_{k}^{L})/\tau) + \sum_{j\notin C_{k}}\exp(sim(z_{i}^{L},m_{j}^{L})/\tau)},
\label{eq:Loss CMCL_L}
\end{equation}
where $\tau$ is the temperature coefficient.
The input $x_{i}^{T}$ from the transverse view shares the same computation method as $x_{i}^{L}$. Thus, the final $L_{CMCL}$ is expressed as the sum of the losses from both views, given as: $L_{CMCL} = L_{CMCL}^{L} + L_{CMCL}^{T}$.

In the aforementioned process, the memory bank $\mathcal{M}$ can be mapped to a hypersphere as shown in Fig.~\ref{fig1}, where $\mathcal{M}$ represents the global feature distribution.
For the features $z^{L}$ and $z^{T}$, the positive pairs correspond to the cluster center from the same class in $\mathcal{M}$, and negative pairs consist of all individual samples in $\mathcal{M}$ that belong to different classes. CMCL forces intra-class features to gather around a global center rather than forming local clusters within a batch, facilitating more stable model optimization, richer and more diverse negative samples help the model learn a more global feature distribution, enhancing the robustness of representation learning.
\subsection{Cascaded Down-Sampling Attention Module}
\label{2.2}
A common multi-view classification approach employs view-specific encoders to extract features and map them to class probabilities via a classifier. Unlike other classification tasks ~\cite{Huang,r13,wang2020auto,huang2023fourier}, carotid plaque regions are relatively small, relying exclusively on high-dimensional features for carotid classification may miss local information. To address this problem, we propose a cascaded DSAM to refine and add the features step by step, achieving smoother feature fusion. 
\begin{figure}[htbp]
    \centering
    \includegraphics[width=1.0\textwidth]{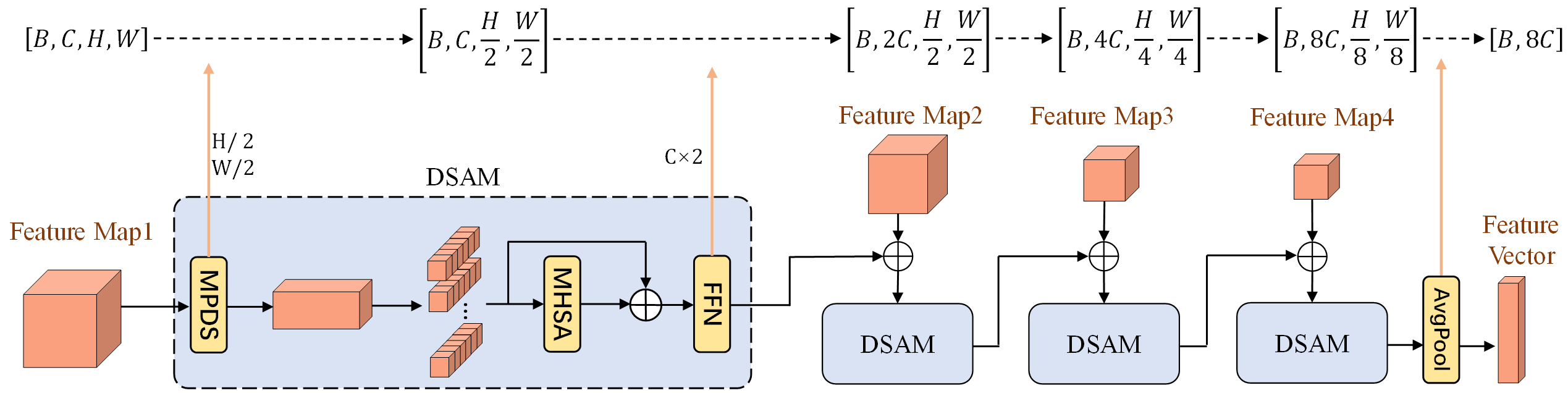}  
    \caption{An illustration of our proposed cascaded down-sampling attention module}%
    \label{fig3}
\end{figure}
As shown in Fig.~\ref{fig3}, four DSAMs are employed to extract and fuse features from different feature maps. A feature map with a shape of $B \times C \times H \times W$ is first fed into the DSAM. It then undergoes a max-pooling down-sampling (MPDS) module to reduce its spatial dimensions. Next, the down-sampled feature is divided into patches along the channel dimension and passes through a multi-head self-attention (MHSA) module. Finally, a feed-forward network (FFN) doubles feature channels after patches are restored to their original spatial shape.
After processing through the DSAM, the first-layer feature map is refined and reshaped to align with the dimensions of the second-layer feature map. These adjusted features are subsequently summed and forwarded to a second DSAM, where an analogous process is applied. The last DSAM output undergoes global average pooling to obtain the feature vectors. The features obtained from all views are then concatenated to form the final fusion feature $z^{F}$.

Through these processes, the texture information from shallow layers is progressively refined by DSAMs, reducing discrepancies between different levels of representation. This mitigates confusion caused by excessive variation in feature scales and allows smoother feature fusion. The shared weight strategy promotes consistent feature extraction across different views, enhancing cross-view feature integration while improving overall feature representation through the enforcement of a unified transformation.
\subsection{Parameter-Free Mixture-of-Experts Weighting Strategy}
\label{2.3}
The MoE model typically consists of a learnable gating network and multiple experts. The gating network dynamically assigns weights to each expert based on the input.
We designed a new gating mechanism that applies weights to experts without extra parameters. 
This enables each expert to focus on specific class information, leading to better category-level feature decoupling. 

For a pair of input samples $x^{L}$ and $x^{T}$, we obtain their representation features $z^{L}$, $z^{T}$ and the fusion feature $z^{F}$. $z^{F}$ is then fed into all experts. For a class-specific expert k, its weight $w_{k}$ is given by:
\begin{equation}
w_{k} = \frac{\exp(sim(z^{cat}, \mu_{k}^{cat}))}{\sum_{i=1}^{K}\exp(sim(z^{cat}, \mu_{i}^{cat}))} 
\label{eq:weight}
\end{equation}
where $z^{cat}=concat(z^{L}, z^{T})$, $\mu_{k}^{cat}=concat(\mu_{k}^{L}, \mu_{k}^{T})$, $\mu_{k}^{L}, \mu_{k}^{T}$ are obtained from Eq.~\eqref{eq:class center}, represent as the class center. Our motivation is as follows: when the feature $z^{cat}$ has a higher similarity to a specific class center $\mu_{k}^{cat}$, the input sample is more likely to belong to that class. The corresponding expert should be assigned a higher weight to learn class-specific features. The parameter-free gate dynamically assigns weights to different experts, allowing them to focus on specific class information. By modeling different classes separately, the experts help reduce feature confusion between classes and improve the model’s discriminative ability. Finally, the output of the MoE is fed into the classifier to obtain the final prediction. The overall loss is formulated as follows.
\begin{equation}
L = L_{CE} + \lambda L_{CMCL}
\label{eq:weight}
\end{equation}
where $L_{CE}$ is the standard Cross-Entropy Loss. $L_{CMCL}$ is the contrastive loss from Eq.~\eqref{eq:Loss CMCL_L} and $\lambda$ is the balancing coefficient. The parameter-free weighting strategy assigns explicit category priors via cosine similarity between input features and the class center. Higher similarity increases the expert’s weight. Unlike trainable gating favors certain experts, parameter-free gating provides stable weight distribution and better guides experts to learn category-specific cues, enhancing interpretability.
\section{Experimental Results}
\textbf{Dataset and Implementations.} We collected a private dataset comprising 1,657 pairs of carotid US images (518 RADS1, 772 RADS2, 367 RADS3-4) from 1,228 patients, with each patient contributing no more than two cases. This study was approved by the local IRB. All images were resized to 224×224 and randomly split into 7:1:2 for training, validation, and testing. Pretrained weights from ImageNet were used for initialization.The CVC-RF was implemented using the PyTorch framework and trained for 100 epochs on an NVIDIA GeForce RTX 4090 GPU, using AdamW optimizer with a learning rate of 1e-4. The model infers in 41.96 ms/case, with 18.73M parameters, 16.50 GFLOPs, demonstrating strong deployment potential.The EMA momentum coefficient $\alpha$, CMCL temperature Coefficient $\tau$, and loss Weight $\lambda$ are set to 0.01, 0.01, and 0.2, respectively. All hyperparameters are determined through experiments on the validation set.
\begin{table}[h!]
    \centering
    \caption{Performance comparison on carotid plaque grading.}
    \renewcommand{\arraystretch}{1.0}
    \resizebox{\textwidth}{!}{
        \begin{tabular}{c|c|c|c|c|c|c|c}
            \toprule
            Method & Acc (\%) & M-Pre (\%) & M-Rec (\%) & M-F1 (\%) & Acc1 (\%) & Acc2 (\%) & Acc3 (\%)  \\ 
            \midrule
            Longitudinal          & 82.41 & 82.20 & 81.31 & 81.74 & 88.57 & 83.23 & 72.00 \\
            Transverse            & 85.43 & 85.41 & 84.99 & 85.10 & 89.52 & 85.16 & 80.00 \\
            \midrule
            Dual-ResRet18
            ~\cite{ResNet18}      & 86.98 & 88.04 & 85.27 & 86.39 & 90.48 & 90.32 & 74.6 \\
            ETMC~\cite{ETMC}      & 88.19 & 87.12 & 87.65 & 87.30 & \textbf{97.14} & 85.81 & 80.00 \\
            DeepGuide
            ~\cite{DeepGuide}     & 88.74 & 90.44 & 86.36 & 87.98 & 89.52 & 94.84 & 74.67 \\
            MVC
            ~\cite{Huang}         & 89.36 & 90.16 & 85.40 & 88.16 & 90.56 & 93.27 & 76.33 \\
            CheXFusion
            ~\cite{chefusion}     & 90.58 & 91.28 & 87.14 & 89.44 & 91.83 & 94.46 & 82.14 \\
            \midrule
            Ours w/o CMCL         & 87.80 & 87.54 & 87.66 & 87.50 & 94.29 & 85.81 & 82.67 \\
            Ours w/o DSAM         & 92.20 & 91.65 & 92.44 & 91.95 & 97.14 & 90.00 & \textbf{90.00} \\
            Ours w/o MoE          & 91.45 & 91.15 & 91.23 & 91.10 & 96.19 & 90.65 & 86.67 \\
            Ours                  & \textbf{93.25} & \textbf{93.80} & \textbf{92.50} & \textbf{93.05} & 94.29 & \textbf{95.16} & 88.00 \\  
            \bottomrule
        \end{tabular}
    }
    \label{tab:results}
\end{table}

\noindent \textbf{Quantitative and Qualitative Analysis.} 
We evaluate the performance of CVC-RF using accuracy (Acc), mean precision (M-Pre), mean recall (M-Rec), mean F1-score (M-F1), and per-class accuracy (Acc1-3). All results were evaluated using paired t-tests with p-values<0.05. Table.~\ref{tab:results} presents a comparison of CVC-RF with single-view methods, multi-view feature fusion~\cite{chefusion,DeepGuide} and weighting approaches ~\cite{ETMC,Huang}, as well as the results of ablation experiments for each module. The experimental results demonstrate that combining both longitudinal and transverse views for carotid plaque grading outperforms single-view approaches. Moreover, our method achieves a significant advantage over other approaches, with Acc3(Minority Class), M-Rec, and M-F1 improving by 4.53\%, 5.36\%, and 3.61\%  compared to the state-of-the-art method~\cite{chefusion}. This highlights the strong advantage of our hierarchical feature refinement strategy over view-wise interaction methods while also helping to mitigate the impact of data imbalance.

\begin{figure}[h!]
    \centering
    \includegraphics[width=1.0\textwidth]{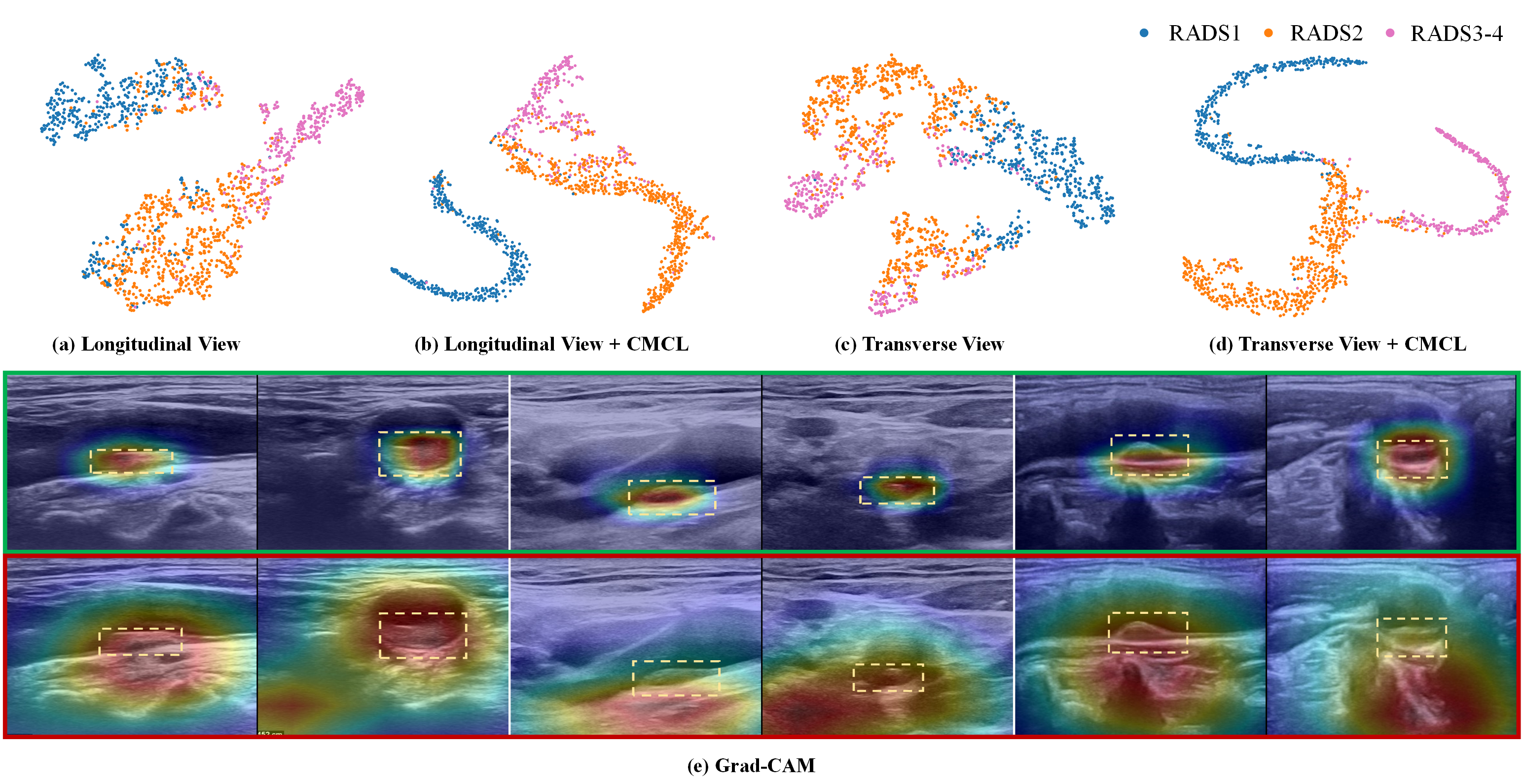}  
    \caption{
    t-SNE and Grad-CAM visualizations are presented. The upper section displays t-SNE results with and without CMCL, and (e) compares Grad-CAM with(Green) and without(Red) DSAM. 
    The yellow box highlights the plaque.
    }
    \label{fig4}
\end{figure}
In the ablation study, we found that the removal of the DSAM or MoE module led to a performance drop, demonstrating their effectiveness. Moreover, ablating CMCL resulted in a significant 5.54\% drop in Acc. This suggests that strong feature representation is essential to establish a solid foundation for subsequent tasks and progressive refinement enhances model performance.  

Fig.~\ref{fig4} presents the t-SNE and Grad-CAM of both views. Without CMCL guidance, features of the same class tend to form multiple small clusters, while class boundaries remain unclear. In contrast, CMCL encourages intra-class features to cluster tightly within the feature space and enforces more distinct decision boundaries between different classes. (e) shows the superiority of DSAM in Grad-CAM on three image pairs, helping the model focus on plaque regions when DSAM is included.

\section{Conclusions}
We propose a novel CVC-RF for multi-view CPG, the first work for CPG based on the plaque-RADS guidelines. By incorporating hierarchical refinement across the ~\textbf{Corpus-level}, ~\textbf{View-level} and ~\textbf{Category-level}, our approach enhances model performance. Additionally, a multi-scale feature fusion strategy is employed to address the challenge of small plaque regions. Experimental results demonstrate the superior performance of our method over existing approaches, validating the effectiveness of multi-level information refinement.

\begin{credits}
\subsubsection{\ackname} This work was supported by the grant from National Natural Science Foundation of China (No.12326619,62171290,82201851), Science and Technology Planning Project of Guangdong Province (No.2023A0505020002), Frontier Technology Development Program of Jiangsu Province (No.BF2024078), Guangxi Province Science Program (No.2024AB17023), Yunnan Major Science and Technology Special Project Program (No.202402AA310052), Yunnan Key Research and Development Program (202503AP140014). Young Talent Program of the Academician Fund (YS-2022-005), Shenzhen Medical Research Fund(C2401015), Shenzhen High-level Hospital Construction Fund(NCRCSZ-2023-003)

\subsubsection{\discintname}
The authors have no competing interests to declare that are relevant to the content of this article.
\end{credits}

\bibliographystyle{splncs04}
\bibliography{reference}

\end{document}